%% file: arxiv.tex
\documentclass{article} 
\usepackage[final]{colm2025_conference}

\usepackage{microtype}
\usepackage{hyperref}
\usepackage{url}
\usepackage{booktabs}
\usepackage{enumitem}
\usepackage{multirow}
\usepackage{wrapfig}

\usepackage{lineno}

\usepackage{amsmath}
\usepackage{amssymb}
\usepackage{amsfonts}
\usepackage{mathtools}
\usepackage{amsthm}
\input{math_commands}

\usepackage{algorithm}
\usepackage[noend]{algpseudocode}

\usepackage[capitalize,noabbrev]{cleveref}

\theoremstyle{plain}

\theoremstyle{definition}

\theoremstyle{remark}

\newcommand{\methodname}{\textsc{Resona}}

\definecolor{darkblue}{rgb}{0, 0, 0.5}
\hypersetup{colorlinks=true, citecolor=darkblue, linkcolor=darkblue, urlcolor=darkblue}

\title{
    \includegraphics[height=15pt]{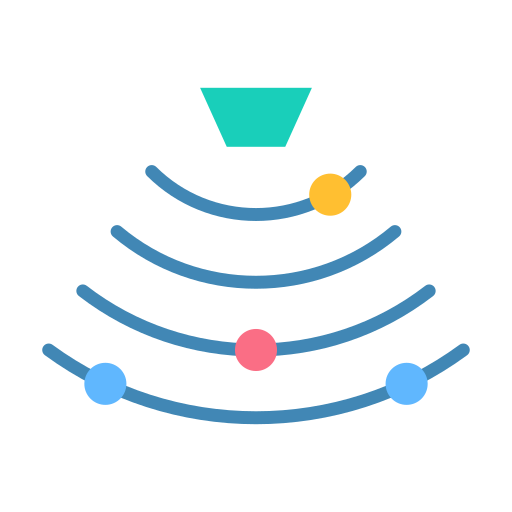}\textsc{Resona}: Improving Context Copying in Linear Recurrence Models with Retrieval
}

\author{\textbf{Xinyu Wang}$^\spadesuit$\thanks{Equal Contribution}
    \ \ 
    \textbf{Linrui Ma}$^\clubsuit$\footnotemark[1] 
    \ \ 
    \textbf{Jerry Huang}$^{\heartsuit\diamondsuit}$\footnotemark[1]
    \ \ 
    \textbf{Peng Lu}$^\heartsuit$
    \\
    \textbf{Prasanna Parthasarathi}$^\clubsuit$
    \ \ 
    \textbf{Xiao-Wen Chang}$^\spadesuit$
    \ \ 
    \textbf{Boxing Chen}$^\clubsuit$
    \ \ 
    \textbf{Yufei Cui}$^\clubsuit$\thanks{Corresponding author: \href{mailto:yufei.cui@huawei.com}{yufei.cui@huawei.com}}
    \\ \\
    \textsuperscript{$\spadesuit$} McGill University \ \textsuperscript{$\clubsuit$} Noah's Ark Lab, Montreal \ \textsuperscript{$\heartsuit$} Universit\'{e} de Montr\'{e}al \ 
    \textsuperscript{$\diamondsuit$} Mila 
}

\begin{document}

\ifcolmsubmission
\linenumbers
\fi

\maketitle

\begin{abstract}
Recent shifts in the space of large language model (LLM) research have shown an increasing focus on novel architectures to compete with prototypical Transformer-based models that have long dominated this space. Linear recurrent models have proven to be a viable competitor due to their computational efficiency. However, such models still demonstrate a sizable gap compared to Transformers in terms of in-context learning among other tasks that require recalling information from a context. In this work, we introduce \methodname, a simple and scalable framework for augmenting linear recurrent models with retrieval. \methodname~augments models with the ability to integrate retrieved information from the provided input context, enabling tailored behavior to diverse task requirements. Experiments on a variety of linear recurrent models demonstrate that \methodname-augmented models observe significant performance gains on a variety of synthetic as well as real-world natural language tasks, highlighting its ability to act as a general purpose method to improve the in-context learning and language modeling abilities of linear recurrent LLMs.
\end{abstract}

\section{Introduction}

Improvements in building state-of-the-art large language models (LLMs)~\citep{gpt4, llama3, qwen2.5, gemma2} through increased scale~\citep{instruction-tuning, scaling} and downstream tuning~\citep{rlhf, alpaca} have enabled them to attain human-level performance on a number of complex tasks. One feature that has enabled this is \textbf{in-context learning}~\citep{gpt3}, where models can use user-provided content to provide a specific response tailed to that example. This relies on the Transformer~\citep{transformers} backbone that underlies many of these models, enabling for models to observe the complete past when generating content.

\begin{figure*}[ht!]
    \centering
    \includegraphics[width=0.95\linewidth]{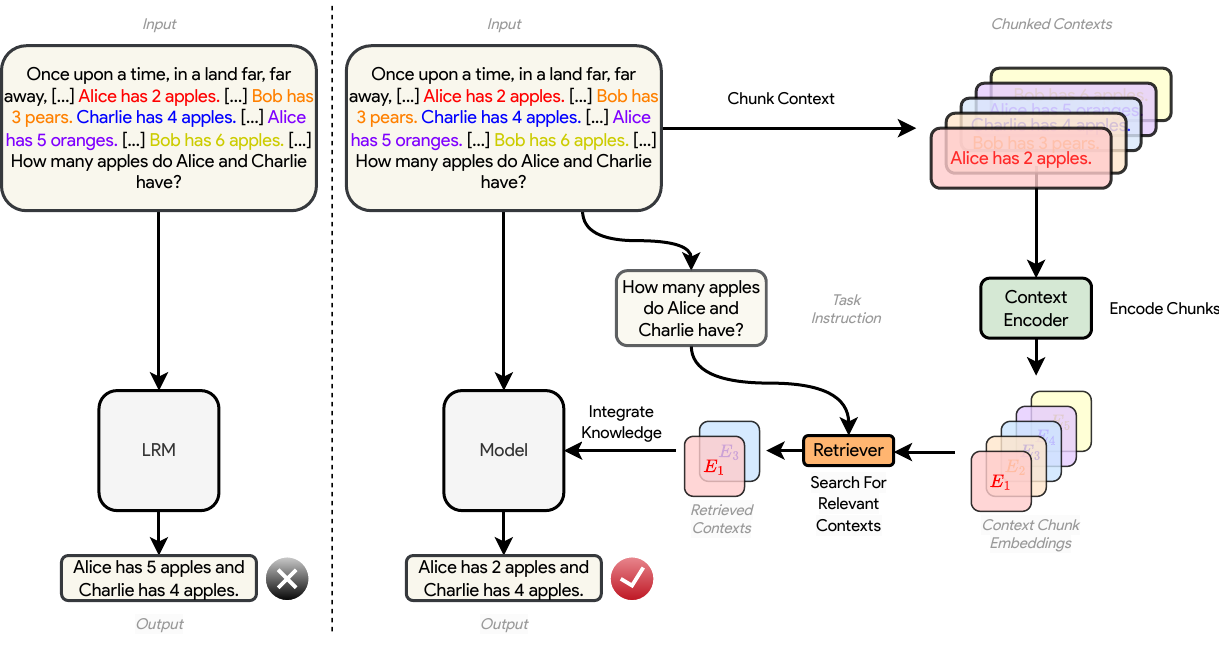}
    \caption{
        Simplified overview of \methodname. The input is separated into the example-specific context along with a task-specific instruction. The context is chunked and encoded, after which a retriever uses the instruction to determine which chunks are relevant to solving the task. The retrieved context is integrated into the model's reasoning for improved response.
    }
    \label{fig:example}
\end{figure*}

Recently, linear recurrent models (LRMs)~\citep{s4, rwkv, lru, mamba, DeltaNet} have emerged as an alternative, aiming to address computational bottlenecks associated with attention mechanisms~\citep{attention}. Unlike Transformers, LRMs do not operate over all previous parts of the input. 
Instead they compress prior context into a unified hidden representation/state with a recurrent structure, similar to original recurrent neural networks (RNNs)~\citep{rnn, lstm, gru, linear-rnn, sclstm, parallel_scan_rnn}, enabling more efficient inference. However, this unified representation introduces an information bottleneck, as it limits the capacity to represent the full range of vocabulary elements within a fixed-size state that does not scale with the combinatorial complexity of token sequences.
This has raised questions about the ability of LRMs to effectively learn from input contexts~\citep{transformers_copying, mamba-icl, mamba-icl2, ssm_pooler} and perform comparably to Transformer-based LLMs in such settings.

Linear methods rely on fixed-size states to address
limitations of traditional attention. A trade-off, however, is that these fixed-size states cannot perfectly preserve all historical information, making exact retrieval challenging. This manifests itself in practice in various tasks such as language modeling; with the key-value associative memory system that underlies such methods, adding new key-value associations leads to accumulating retrieval errors that hinder performance. These errors may build up in a variety of data-independent~\citep{s4, lru} or dependent~\citep{gla, mamba, eagle_finch, gsa, xlstm} manners. While recent works have proposed strategies  to  mitigate such errors~\citep{DeltaNet, gatedDeltaNet, ttt}, they retain a fixed-size hidden state that remains a fundamental constraint.

In an effort to bridge this gap, we propose \methodname, a retrieval-based method designed to improve context-based learning in LRMs (\cref{fig:example}). By introducing a retrieval mechanism that facilitates information flow from the context, \methodname 
mitigates the hidden state bottleneck and enables more effective in-context learning. Specifically, we augment LRM layers within the backbone model with a contextual search component. The input context is first chunked into passages, which are then retrieved based on the current LRM state. A knowledge integration module subsequently incorporates the retrieved passages into the model's output by directly modifying its representation. This architecture allows previous context to bypass the fixed-size hidden state constraint, improving information flow from context to generation.These processes (chunking, retrieval, and integration) are parallelized with the main LRM, ensuring easy adaptation and improved in-context learning across a variety of models.

Empirical results on a range of representative tasks, spanning both synthetic and real-world data, demonstrate that \methodname~significantly improves the ability of LRMs to utilize context-specific information with minimal or no latency overhead. 
We evaluate \methodname~on a diverse set of tasks, including synthetic retrieval and recall tasks, language modeling and question-answering tasks, across multiple scenarios such as pre-training and direct fine-tuning. Our analysis demonstrates the effectiveness of using \methodname~for overall performance improvements as well as test-time model customizations, such as balancing the trade-off between efficiency and performance.

\section{Related Works}

\paragraph{Linear Recurrent Models.} Despite vast improvements in building language models that solve real-world natural language tasks since the introduction of the Transformer~\citep{transformers}, significant concerns remain about their efficiency and scalability. While this has spurred interest in rendering them more efficient~\citep{linear-attention, fla, gla}, LRMs~\citep{s4, lru, hgrn, mamba, mamba2, rwkv, retnet, regla, DeltaNet} have grown as a popular alternative due to highly efficient inference costs compared to attention-based alternatives while retaining the ability to be train on elements of a sequence in parallel, an issue with traditional recurrent models. Further attempts at leveraging advantages from both paradigms~\citep{jamba, griffin-hawk, hymba} have also garnered interest, leading to further exploration of similar models.

\paragraph{Retrieval-Augmented Generation (RAG).} RAG-based methods augment the input of an LM with passages retrieved from an outside source~\citep{rag, rag2}. Such methods have significantly improved performance on knowledge-intensive tasks, where it is difficult to store the information required for strong performance explicitly within the model’s parametric knowledge~\citep{parameters-kb}. Further improvements have come under the form of improved filtering of retrieved passages~\citep{self-rag}, robustness to irrelevant passages~\citep{retrieval-robustness, recomp} or tuning of more components~\citep{ra-dit}. However, RAG is not directly applicable to learning from contexts, as such methods do not search within the input-specific query but rather from an external database, leading to a critical failure point of such methods.

\paragraph{Linear Recurrent Models and Context Usage.} 
Despite their practical benefits, questions have arose regarding the ability of LRMs to learn from input contexts~\citep{icll}. \citet{transformers_copying} show that they struggle to directly copy information from contexts due to their fixed-sized latent state. \citet{mamba-icl} further show that they can struggle at retrieval-based~\citep{zoology_mqar} in-context learning, only solving such tasks through the addition of attention. Such observations have extended to real-world data, where LRMs have been shown to exhibit similar difficulties as Transformer LLMs~\citep{resonance_rope, extrapolation, litm, hallucination} for long contexts~\citep{sled-niah, ruler}. Accordingly, we introduce \methodname~as a potential solution that provides additional information flow paths from the context to the generated input, enabling better utilization of the context for problem-solving.

\paragraph{Memory-enhanced Transformers.} Due to the quadratic complexity of self-attention with respect to the length of a sequence, Transformer models face significant computational challenges when processing long inputs. Numerous approaches have been proposed to enhance Transformers for long sequential data, both to reduce the time/space complexity of the models as well as to improve performance. \citet{Transformer-XL, infini-attn} segment long inputs into shorter sequences and process them recurrently. \citet{Landmark-Attn} append landmark tokens to represent each block of input and uses group attention to select relevant information. \citet{retro} enhance Transformer performance with external data by leveraging a separated retriever module. \citet{Expansion-Span} develop a Span-Expanded Attention for the hybridized attention model to retrieve the most relevant block and integrate it with the recent context for attention computation. However, it remains unclear whether and how retrieval-based modules can enhance the performance of generalized linear recurrent models such as GLA~\citep{gla_mao, gla, regla}, Mamba~\citep{mamba, mamba2}, RWKV~\citep{rwkv}, and DeltaNet~\citep{deltarule, DeltaNet}.

\section{\methodname}

We introduce \methodname~(\cref{fig:echo} and \cref{alg:echo}) as a framework to enhance the context-copying ability of LRMs through retrieval, without sacrificing original performance and versatility. Our end-to-end training enables models to utilize the context as a retrieval base from which information can be extracted.
This helps the model to first overcome the fixed-size latent space bottleneck by integrating information from the context directly into the hidden state. This is in contrast to traditional LRMs, where the information from the context must flow through the hidden state.

\begin{minipage}{0.45\textwidth}
\begin{algorithm}[H]
\small
\caption{\methodname~Algorithm.}\label{alg:echo}
    \begin{algorithmic}[1]
        \Require Model $M$
        \State {\bf Input:} Input $\vs\in\mathbb{R}^{T}$ and LRM hidden states $\tH \in\mathbb{R}^{L\times T\times H}$, 
        \State {\bf Output:} $\mY\in\mathbb{R}^{T\times D}$
        \State Embed sequence $\vs$ into $\mX\in\mathbb{R}^{T\times D}$.
        \For {$l \gets 1$ to $L$}                     \State $\mH \gets \tH_l$ \Comment{Hidden state of layer $l$}
            \If{Layer $l$ is an \methodname~layer}
                \State $(\mH, \mY^{m})=M_l^{\text{LRM}}\left(\mX, \mH\right)$ 
                \State \Comment{LRM Output}
                \State $\mM = M_l^\text{C-and-S}(\mX, \mH)$ 
                \State \Comment{Chunk-and-Search}
                \State $\mY^{r} = M_l^\text{KI}\left(\mX, \mH, \mM\right)$ 
                \State \Comment{Knowledge Integration}
                \State $\alpha = f\left(\mX\right)$
                \State $\mY=\alpha\cdot\mY^m+(1-\alpha)\mY^r$
            \Else
                \State $(\mH, \mY)=M_l\left(\mX, \mH\right)$
            \EndIf 
            \State $\mX=\mY$
        \EndFor
        \State \textbf{return $\mY$}
    \end{algorithmic}
\end{algorithm}
\end{minipage}
\hfill
\begin{minipage}{0.53\textwidth}
\begin{figure}[H]
    \centering
    \includegraphics[width=\linewidth]{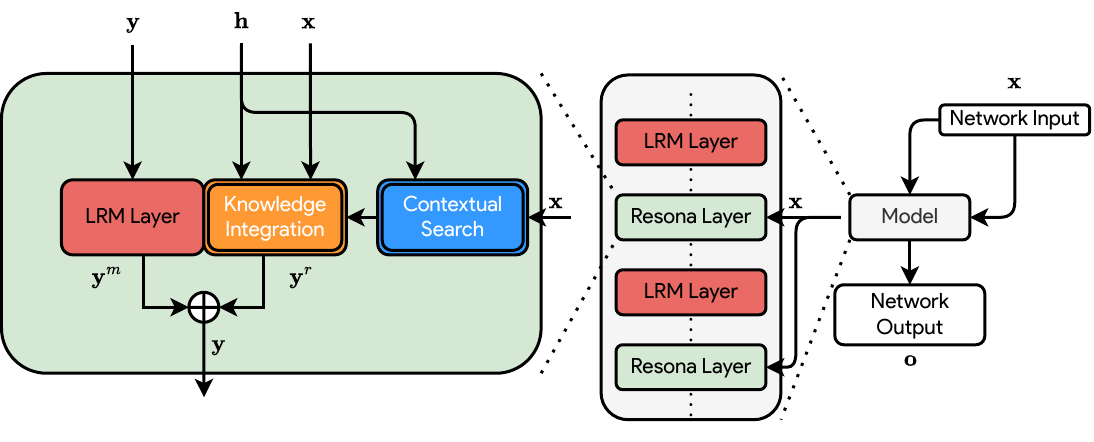}
    \caption{Primary components of \methodname: 1) \textbf{Contextual Search} which searches the context for relevant information and 2) \textbf{Knowledge Integration} which re-integrates the retrieved information into the model state. These enable information from the original network input to flow to arbitrary depths in the model, overcoming information decay within the model.}
    \label{fig:echo}
\end{figure}
\end{minipage}

\subsection{Problem Formulation and Overview}

Let $\mathcal{V}$ be a vocabulary, i.e. a set of discrete elements, of size $\left|\mathcal{V}\right|$.
A model $M$ applies a function $f: \mathcal{V}^* \to \mathcal{V}^*$, taking as input a sequence of tokens from the vocabulary (of arbitrary length) while outputting a sequence of tokens from the vocabulary. We denote the input sequence as $\vx=[x_1, \dots, x_T$], which we refer to as the model’s prompt.  
The corresponding output sequence $\vy=f(\vx)$ is referred to as the model’s answer or generated response. 
Furthermore, a sequence-to-token mapping is a function $g: \mathcal{V}^* \to \mathcal{V}$ used to define $f$ through auto-regressive inference.  Specifically, given an input sequence $\vx \in \mathcal{V}^*$, the output tokens are generated one at a time using the recurrence:
$x_{i+j} = g(x_1, \dots, x_{i+j-1})$ and $f\left(\vx_{1:i}\right) = (x_{i+1}, x_{i+2}, \dots)$, where $1\leq i\leq T$ and $j \in\mathbb{N}$.

An LRM is defined by a state update rule $u: \mathcal{S} \times \mathcal{V} \to \mathcal{S}$ and an output function $r: \mathcal{S} \to \mathcal{V}$, where $\mathcal{S}$ is a finite set of states and a state is a representation of the system after processing a sequence from $\mathcal{V}^*$. Let $s_0 \in \mathcal{S}$ be some initial state. Given some sequence $\vx$ of length $L$, the state of the model at iteration $i$ is denoted by $S_i(x_1, \dots, x_i)$ and the output token is denoted by $R_i(x_1, \dots, x_i)$. These are defined recursively as:
\begin{enumerate}[parsep=0pt,label=\arabic*)]
    \item $S_0\left(\emptyset\right) = s_0$,
    \item $\vh_i=S_i(x_1, \dots, x_i) = u\left(S_{i-1}(x_1, \dots, x_{i-1}), x_i\right)$,
    \item $y_i=R_i(x_1, \dots, x_i) = r\left(S_{i}(x_1, \dots, x_{i})\right)$.
\end{enumerate}

Information from $\vx$ flows to $\vy$ through a state $\vh\in\mathbb{R}^{d_h}$ where $d_h$ is fixed and finite. Thus for increasing sequence length or information dense settings, LRMs can struggle from the limited size of $\vh$. 

We observe that the LRM is directly limited by the size of its hidden state, which can be insufficient for modeling problems with many possible states, unless the hidden size grows with the size of the possible set of states. \methodname~introduces two flexible components to overcome this constraint without sacrificing the primary benefits of LRMs (namely parallel training and inference time efficiency): 1) a \textbf{contextual search} operation that operates on the input to retrieve context-specific information and 2) a \textbf{knowledge integration} component that mixes the retrieved information with the LRM output.

\begin{figure}[ht]
\begin{minipage}{0.49\textwidth}
    \begin{figure}[H]
    \centering
    \includegraphics[width=\linewidth]{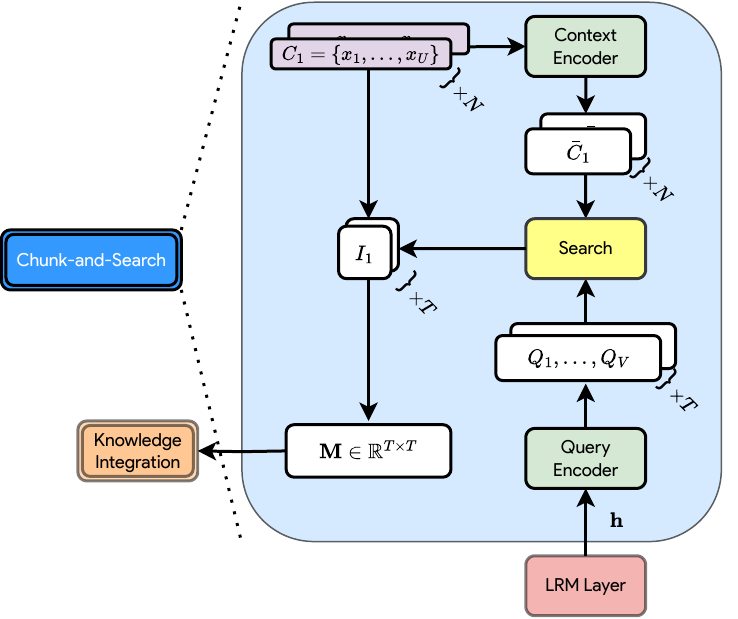}
    \end{figure}
\end{minipage}
\hfill
\begin{minipage}{0.49\textwidth}
\begin{algorithm}[H]
\small
    \caption{Chunk-and-Search Algorithm.}\label{alg:chunk-and-search}
    \begin{algorithmic}[1]
        \Require Context and Query Encoders $\mathcal{C}$, $\mathcal{Q}$
        \State {\bf Input:} Input $\mX\in\mathbb{R}^{T\times D}$ and LRM hidden state $\mH \in\mathbb{R}^{T\times H}$. 
        \State {\bf Output:} Attention mask $\mM \in\mathbb{R}^{T\times T}$
        \State Chunk $\mX$ into $\tX'\in\mathbb{R}^{N\times U\times D}$
        \State Use $\mathcal{C}$ to encode each context chunk $\{\mX'_i\}_{i=1}^N\in\tX'$ into context embeddings $\bar{\mC}\in\mathbb{R}^{N\times E}$
        \State Use $\mathcal{Q}$ to encode $\mH$ into query embeddings $\bar{\mQ}\in\mathbb{R}^{T\times E}$.
        \State Compute chunk index sets for each $\bar{\mQ}_j$: $\{I_j\}_{j=1}^T=\{\text{Top-}k(\bar{\mQ}_{j}, \bar{\mC})\}_{j=1}^T$
        \State With $\{I_j\}_{j=1}^T$, compute a mask $\mM\in\mathbb{R}^{T\times T}$ such that $\mM_{ji} = 1 \iff \left(i\in\{I_j\}\right)$.

    \end{algorithmic}
\end{algorithm}
\end{minipage}

    \caption{A breakdown of our Chunk-and-Search implementation. The initial input context is chunked while the hidden state of the LRM layer is used as a query. Corresponding indices are retrieved for each query, creating a mask that is used for Knowledge Integration.}
    \label{fig:chunk-and-search}
\end{figure}

\paragraph{Contextual Search.} In order to retrieve relevant context, \methodname~implements contextual search as a chunk-and-search mechanism~(\cref{alg:chunk-and-search}). The initial input $\mX$ is first split into $N$ chunks, each consisting of $U$ tokens, to create $\tX'\in\mathbb{R}^{N\times U\times D}$, where $D$ is the model input dimension. First, each of these chunks $C_i$ is encoded using a context encoder $\mathcal{C}$ into a context embedding $\bar{C}_i$. Simultaneously, the hidden state  $\mH\in\mathbb{R}^{T\times H}$ of the adjacent linear-recurrent layer, is used to encode a number of queries into query embeddings $\bar{Q}_{1:T}$ using a query encoder $\mathcal{Q}$. For each query, we search for the top-$k$ similar contexts using a cosine-similarity search, which produces chunk indices that we can then use to retrieve the relevant input token positions. These are used to create a mask, which is passed to the \textbf{Knowledge-Integration} module.

\begin{figure}[ht]
\begin{minipage}{0.49\textwidth}
    \begin{figure}[H]
    \centering
    \includegraphics[width=\linewidth]{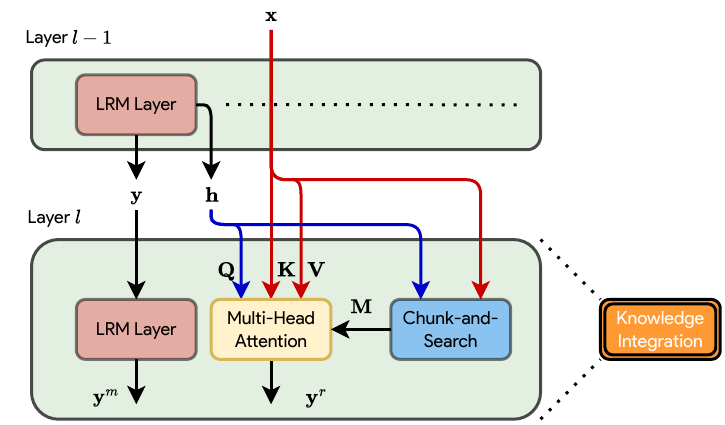}
    \end{figure}
\end{minipage}
\hfill
\begin{minipage}{0.48\textwidth}
\begin{algorithm}[H]
\small
    \caption{Knowledge Integration.}\label{alg:knowledge-integration}
    \begin{algorithmic}[1]
    \Require Attention weights $W_Q, W_K, W_V$ and output weights $W_{\text{out}}$
    \State {\bf Input:} Input $\mX\in\mathbb{R}^{T\times D}$ and LRM hidden state $\vh\in\mathbb{R}^{T\times H}$, mask $\mM \in\mathbb{R}^{T\times T}$
    \State {\bf Output:} $\mY^r\in\mathbb{R}^{T\times D}$
    \State From $\vh$, compute queries $\mQ\in\mathbb{R}^{T\times d_k}$ using $W_Q$. In parallel, compute $\mK, \mV \in \mathbb{R}^{T\times d_k}$ from $\mX$ with $W_K$, $W_V$.
    \State Compute multi-head attention output $\mO =\mathrm{CrossAttn}\left(\mQ, \mK, \mV, \mM\right)$, where $\mO \in\mathbb{R}^{T\times d_k}$.
    \State Project $\mO$ using $W_{\text{out}}$ into $\mY^r\in\mathbb{R}^{T\times D}$.
    \end{algorithmic}
\end{algorithm}
\end{minipage}
    \caption{Knowledge Integration portion of \methodname. The hidden state is used as a query while the initial network input is used as the query and keys. A mask constructed from the contextual search is used to determine which information needs to be mixed back into the LRM representation, which is then used to compute an attention output that is integrated into the adjacent LRM output.}
    \label{fig:knowledge-integration}
\end{figure}

\paragraph{Knowledge Integration.} To integrate knowledge from the retrieved chunks, \methodname~does as follows~(\cref{fig:knowledge-integration} and \cref{alg:knowledge-integration}). The knowledge integration module is a cross-attention module that can directly integrate information from the initial embedding into the LRM layer representation. To do so, the queries $\mQ$ are directly computed from the hidden state of the prior LRM layer\footnote{In the event that the first layer is augmented with \methodname, the queries are generated directly from the initial embeddings.}, while the keys $\mK$ and values $\mV$ are computed directly from the input embeddings $\mX$ that directly follow after the initial embedding matrix $\mE$. Within the cross-attention module, we use a mask computed from our \textbf{Chunk-and-Search} implementation of the contextual search. This ensures that the cross attention module can mix in only the most relevant information from the input back into the cross-attention module, producing an output $\mY^r\in\mathbb{R}^{T\times D}$, which is then integrated with the output of the adjacent LRM layer $\mY^m$, computed as

\[
\mY = \alpha\cdot \mY^m+(1-\alpha)\cdot \mY^r.
\] 
$\alpha$ can be computed on an input-dependent basis for each element of $\mY$ or can alternatively be set as fixed hyper-parameter value. Because only the chunk-and-search design, each query attends to at most $kU$ elements from the initial input in $k$ contiguous blocks, enabling us to compute $\mY^r$ efficiently using existing sparse attention mechanisms. For simplicity, we maintain the use of a fixed constant $\alpha$ for the experiments that follow.

\subsection{Training and Inference}

When training \methodname, an important consideration is the auto-regressive nature of the model that therefore requires a causal mask. Specific to our implementation, to maintain this during the parallel nature of training, we ensure that for the query at index $i$, only chunks that solely contain information prior to this position in the input are considered within the \textbf{Chunk-and-Search} process. This ensures that the mask $\mM$ allows no information from a given token to affect the representation of those ahead of it in the sequence.

During inference, tokens are dynamically chunked based on a predefined size and embedded into a chunked cache in parallel with the main model's embedding, introducing no extra latency. For long prompts, chunking is integrated into the pre-filling stage, aligning with the token embedding pipeline and minimizing computation overhead.

\section{Experiments, Results and Analysis}\label{sec:results}

\subsection{Tasks and Datasets}\label{subsec:tasks}

To test our method, we evaluate \methodname~on both a number of synthetic benchmarks as well as real-world language benchmarks. In \cref{sec:results}, we explain the experimental setting as well as evaluation methods for each. For each setting, we report a standard baseline where a backbone model is not augmented with \methodname. These baselines vary based on which backbones are capable of adequately learning the task without \methodname. 

\subsection{Main Results}\label{subsec:main_results}
\begin{figure*}[t!]
    \centering
    \includegraphics[width=0.75\linewidth]{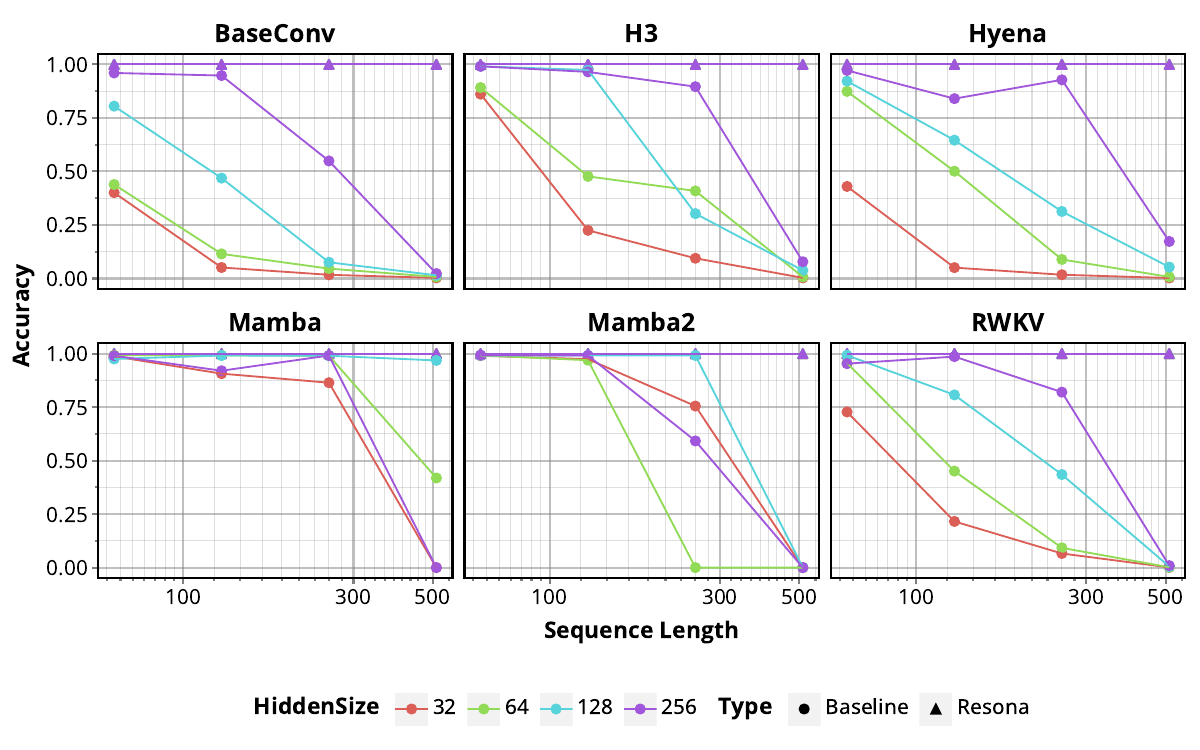}
    \vspace{-1.5\baselineskip}
    \caption{Results on \textsc{MQAR} tasks on varying sequence lengths. Baseline models remain limited in their ability to model increasingly long sequences even with increasing hidden size, whereas augmentation with \methodname~perfect performance on arbitrary lengths.}
    \label{fig:mqar}
\end{figure*}

\begin{table*}[t!]
    \centering
    \caption{Performance on synthetic \textsc{MAD}~\citep{madlab} tasks. The best result for each metric is highlighted in bold. \methodname~consistently boosts retrieval performance and shows gains even on compression and memorization tasks.}
    \resizebox{\linewidth}{!}{
        \begin{tabular}{l|lllllll}
        \toprule
        \textbf{Model} & \textbf{Comp.} & \textbf{ICR} & \textbf{Noisy ICR} & \textbf{Fuzzy ICR} & \textbf{SC} & \textbf{Mem.} & \textbf{Average} \\
        \midrule
        \midrule
        Transformer & 53.4 & 96.7 & 99.7 & 69.4& 98.7 & 89.4 & 84.1 \\
        \midrule
        {Mamba}  & \textbf{38.3} & 76.7 & 74.9 & 9.3 & 33.2 & 88.5 & 53.5 \\
        +\methodname  
        & 38.2 \textcolor{red}{\tiny ($\downarrow$ 0.1)} 
        & \textbf{99.9} \textcolor{teal}{\tiny ($\uparrow$ 23.2)} 
        & \textbf{100.0} \textcolor{teal}{\tiny ($\uparrow$ 25.1)} 
        & \textbf{63.4} \textcolor{teal}{\tiny ($\uparrow$ 54.1)} 
        & \textbf{42.7} \textcolor{teal}{\tiny ($\uparrow$ 9.5)} 
        & \textbf{88.8} \textcolor{teal}{\tiny ($\uparrow$ 0.3)} 
        & \textbf{72.1} \textcolor{teal}{\tiny ($\uparrow$ 18.6)} \\
        \midrule
        {Mamba2} & 43.6 & 96.4 & 96.7 & 21.1 & 93.3 & 86.9 & 73.0 \\
        +\methodname 
        & \textbf{46.6} \textcolor{teal}{\tiny ($\uparrow$ 3.0)} 
        & \textbf{100.0} \textcolor{teal}{\tiny ($\uparrow$ 3.6)} 
        & \textbf{100.0} \textcolor{teal}{\tiny ($\uparrow$ 3.3)} 
        & \textbf{62.9} \textcolor{teal}{\tiny ($\uparrow$ 41.8)} 
        & \textbf{93.6} \textcolor{teal}{\tiny ($\uparrow$ 0.3)} 
        & \textbf{88.1} \textcolor{teal}{\tiny ($\uparrow$ 1.2)} 
        & \textbf{81.9} \textcolor{teal}{\tiny ($\uparrow$ 8.9)} \\
        \midrule
        {RWKV5} & 36.8 & 96.4 & 96.6 & 12.1 & 52.7 & 55.0 & 58.3 \\
        +\methodname  
        & \textbf{40.4} \textcolor{teal}{\tiny ($\uparrow$ 3.6)} 
        & \textbf{99.7} \textcolor{teal}{\tiny ($\uparrow$ 3.3)} 
        & \textbf{99.8} \textcolor{teal}{\tiny ($\uparrow$ 3.2)} 
        & \textbf{59.7} \textcolor{teal}{\tiny ($\uparrow$ 47.6)} 
        & \textbf{58.0} \textcolor{teal}{\tiny ($\uparrow$ 5.3)} 
        & \textbf{70.6} \textcolor{teal}{\tiny ($\uparrow$ 15.6)} 
        & \textbf{71.5} \textcolor{teal}{\tiny ($\uparrow$ 13.2)} \\
        \midrule
        {Hyena} & 42.2 & 79.3 & 77.4 & 9.96 & 72.8 & 88.9 & 61.7 \\
        +\methodname 
        & \textbf{42.6} \textcolor{teal}{\tiny ($\uparrow$ 0.4)} 
        & \textbf{99.9} \textcolor{teal}{\tiny ($\uparrow$ 20.6)} 
        & \textbf{99.9} \textcolor{teal}{\tiny ($\uparrow$ 22.5)} 
        & \textbf{66.2} \textcolor{teal}{\tiny ($\uparrow$ 56.2)} 
        & \textbf{74.3} \textcolor{teal}{\tiny ($\uparrow$ 1.5)} 
        & \textbf{89.0} \textcolor{teal}{\tiny ($\uparrow$ 0.1)} 
        & \textbf{78.7} \textcolor{teal}{\tiny ($\uparrow$ 17.0)} \\
        \bottomrule
        \end{tabular}
    }
    \label{tab:madlab}
\end{table*}

\subsubsection{Results on Synthetic Benchmarks.} 
We first evaluate on synthetic benchmarks, namely multi-query associative recall (\textsc{MQAR})~\citep{zoology_mqar} and the Mechanistic Architecture Design (\textsc{MAD}) suite of tasks~\citep{madlab}. For each, we report accuracy on a held-out test set, where a correct answer requires the entire output is being correctly predicted. We initialize models from scratch and train them on the task of interest, in particular a 4 layer model with a vocabulary size of 8192. Each model uses a hidden size of 128 and a context chunk size of 64 for those augmented with \methodname. Models are trained using 20$\mathsf{K}$ and evaluated on 1$\mathsf{K}$ examples.

\cref{fig:mqar} and \cref{tab:madlab} demonstrate that \methodname~augmentations improves performance across all baselines, some by wide margins. Baseline models are often able to perfectly solve \textsc{MQAR} for shorter sequence lengths or a smaller number of KV-pairs, but fail catastrophically upon increasing either of these values. \methodname meanwhile retains near perfect accuracy even after these values. Similarly, models can struggle at specific tasks within the \textsc{MAD} suite, but \methodname~achieves a significant improvement in performance. On tasks in which the base LRM models show strong performance, no degradation is observed through the addition of \methodname, highlighting its flexibility. We observe that this is consistent across multiple models~\citep{hyena, H3}, which are incapable of learning on some of the simpler settings but upon the addition of \methodname~layers are able of maintaining nearly perfect accuracy for arbitrarily long sequences, showcasing its specific benefits in context-recall intensive settings.

\begin{wraptable}{R}{0.4\linewidth}
    \centering
    \vspace{-\baselineskip}
    \caption{Comparison of pre-training perplexity (PPL) on WikiText-103 across base LRM architectures, their parameter-aligned variants, and \methodname-enhanced versions.}
    \resizebox{\linewidth}{!}{
        \begin{tabular}{lc|c}
            \toprule
            \textbf{Model} & \textbf{Param} & \textbf{PPL} \\
            \midrule
            \midrule
            {GLA} & 131M & 14.265  \\
            GLA (SP) & 142M & 14.223  \\
            +\methodname & 145M & \textbf{13.892} \\
            \midrule
            {DeltaNet} & 131M & 13.044  \\
            {DeltaNet} (SP) & 142M & 12.946  \\
            +\methodname & 145M & \textbf{12.541} \\
            \midrule
            {RetNet} & 132M & 15.471  \\
            {RetNet} (SP) & 146M & 15.431 \\
            +\methodname & 142M & \textbf{14.742} \\
            \midrule
            {Mamba} & 140M & 16.261  \\
            {Mamba} (SP) & 157M & 16.173 \\
            +\methodname & 154M & \textbf{15.943} \\
            \midrule
            {Hymba(64)} & 133M & 16.688 \\
            +\methodname (SP) & 135M & \textbf{15.887} \\
            \bottomrule
        \end{tabular}
    }
    \label{tab:pretraining}
\end{wraptable}

\subsubsection{Language Modeling}

To assess language modeling, we compare a baseline models with one augmented with \methodname~layers. Here, we train on the \textsc{WikiText-103} dataset~\citep{merity2016pointer}. In order to train models augmented with \methodname, we also augment the dataset. Specifically, we first consolidate all samples from the same Wikipedia entry into single sample, eliminating excessively short title lines or empty lines. We then use a \texttt{LLaMA3.1-70B}~\citep{llama3} model to augment each sample such that we can make use of the \methodname~retrieval mechanism. We then conduct the Chunk-and-Search process offline to create masks, in order to save computation during training.
To account for the additional parameters introduced by \methodname, we present results in the baseline settings for two version, one where each LRM layer matches exactly that of the augmented model as well as a version where the hidden size of the layer has been increased to match the parameter count of the augmented counterpart.
Results in \cref{tab:pretraining} demonstrates that integrating \methodname~consistently achieves lower perplexity than baseline counterparts and their parameter-aligned variants, highlighting its applicability for language modeling. Notably, modifying Hymba's sliding window mask to our retrieval-based mask significantly improves performance. Furthermore, results on short-context tasks (\cref{tab:pretraining-detail}) demonstrate no performance degradation, suggesting the ability to maintain performance on tasks that are not recall intensive.

\subsubsection{Direct Fine-Tuning} 

To understand how well the addition of \methodname~modules can improve performance on context-dependent tasks, we make use of pre-trained models in which we insert \methodname~layers. The models are then fine-tuned directly, as described in \cref{subsec:tasks}, and we record performance using task-specific metrics. Due to some computational limitations, we provide results only for models where the baseline model is capable of performance above random on all benchmarks. In these settings, we choose 3 layers of the model to augment with \methodname. The models are then tuned using the corresponding training dataset of the task, before being tested on a held-out test set.

Under this direct-fine-tuning setting, we evaluate on question answering benchmarks including \textsc{NarrativeQA}~\citep{narrativeqaa}, the Conversational Question Answering (\textsc{CoQA}) challenge~\citep{coqa} and \textsc{TriviaQA}~\citep{triviaqa}. For each task, we report results in terms of BLEU, ROUGE-L, Meteor~\citep{meteor} and F1 scores. \cref{tab:qa} presents these results, where we can observe improvements on both pre-trained {Mamba} and Hymba (with a sliding window of 256) models through the addition of \methodname~layers. This is particularly evident with improvements across all metrics for all datasets, showing the general benefits that \methodname~provides towards better context-dependent reasoning skills.

\begin{table*}[t!]
    \centering
    \caption{Results on QA benchmarks, where augmentation with \methodname~improves performance over all evaluation metrics. \textbf{Hymba} indicates \methodname~is added as a third branch with both the Hymba's original linear and attention branches, while \textbf{Hymba(256)} indicates where we reduce the window size of Hymba's sliding window attention from 1024 to 256, in order to create a fair comparison with our chunk size of 256.}
    \resizebox{\linewidth}{!}{
        \begin{tabular}{l|cccc|cccc|cccc}
            \toprule
            \multirow{2}{*}{\textbf{Model}} & \multicolumn{4}{c|}{\textsc{TriviaQA}} & \multicolumn{4}{c|}{\textsc{CoQA}} & \multicolumn{4}{c}{\textsc{NarrativeQA}} \\
            & BLEU & Rouge-L & Meteor & F1 & BLEU & Rouge-L & Meteor & F1 & BLEU & Rouge-L & Meteor & F1 \\
            \midrule
            \midrule
            {Mamba} & 28.3 & 66.0 & 44.8 & 40.2 & 35.5 & 73.0 & 49.3 & 60.6 & 12.6 & 47.0 & 34.7 & 34.7 \\
            +\methodname & \textbf{30.7} & \textbf{68.0} & \textbf{45.6} & \textbf{41.8} & \textbf{44.3} & \textbf{75.0 }& \textbf{52.2} & \textbf{61.2 }& \textbf{13.2} & \textbf{50.0} & \textbf{36.9} & \textbf{36.2} \\
            \midrule
            {Hymba(256)} & 15.3 & 58.0 & 41.4 & 34.3 & 31.8 & 62.0 & 42.1 & 50.0 & 10.7 & 39.0 & 29.1 & 28.4 \\
            
            \textsc{\quad w/\textsc{Retrieval}} & 2.9 & 57.0 & 39.0 & 33.9 & 4.6 & 52.0 & 38.2 & 41.3 & 3.9 & 36.0 & 30.8 & 27.6 \\
            \textsc{\quad w/\textsc{RAG}} & 12.9 & \underline{66.0} & \underline{46.0} & 38.8 & 23.2 & 67.0 & 47.9 & 54.5  & 12.5 & \underline{49.0} & \underline{38.4} & \underline{35.2} \\
            +\methodname & \textbf{25.6} & \textbf{61.0} & \textbf{43.4} & \textbf{41.5} & \textbf{36.5} & \textbf{69.0} & \textbf{48.8} & \textbf{57.2} & \textbf{13.1} & \textbf{46.0} & \textbf{33.9} & \textbf{35.1} \\
            \midrule
            {Hymba} & 16.5 & 64.0 & 44.8 & 36.9 & 40.0 & 77.0 & 53.3 & 63.1 & 20.5 & 59.0 & 44.2 & 43.1 \\
            +\methodname & \textbf{29.3} & \textbf{69.0} & \textbf{48.4} & \textbf{45.4} & \textbf{51.7} & \textbf{82.0} & \textbf{59.3} & \textbf{68.9} & \textbf{20.9} & \textbf{60.0} & \textbf{45.2} & \textbf{44.0} \\
            \bottomrule
        \end{tabular}
    }
    \vspace{-\baselineskip}
    \label{tab:qa}
\end{table*}

\section{Analysis and Discussion}

\paragraph{\methodname~ vs. RAG-like methods.} 
A natural comparison to make with \methodname~is typical RAG-like methods, which append the retrieved passages directly to the input prior to applying the model. To compare with such methods, we design the following methods with \methodname: \textbf{1)} Using the input context directly as the data-store from which a pre-trained retriever can retrieve passages directly that are used as context (\textsc{Retrieval}) and \textbf{2)} we remove the Knowledge Integration component and instead directly append all (decoded) retrieved passages from the Chunk-and-Search procedure to the initial input ({RAG}). 
\begin{wraptable}{R}{0.5\textwidth}
    \centering
    \vspace{-\baselineskip}
    \caption{A comparison of \methodname~and Hymba~\citep{hymba}. Three consecutive layers are un-freezed for fine-tuning. In the \methodname-augmented setup, the attention branch of the selected layers is modified to the \methodname~retrieval mechanism. A sliding window size of 256 is used while the corresponding number indicates the start index of the replaced layers.}
    \label{tab:qa-multidepth}
    \resizebox{\linewidth}{!}{
        \begin{tabular}{lc|cccc}
            \toprule
            \multirow{2}{*}{\textbf{Model}} & \multirow{2}{*}{\textbf{Layer}} & \multicolumn{4}{c}{\textsc{\textbf{CoQA}}} \\
            & & BLEU & Rouge-L & Meteor & F1 \\
            \midrule
            \midrule
            {Hymba} & \multirow{2}{*}{0} & 31.8 & 62.0 & 42.1 & 50.0\\
            +\methodname & & \textbf{36.5} & \textbf{69.0} & \textbf{48.8} & \textbf{57.2} \\
            \midrule
            {Hymba} & \multirow{2}{*}{8} & 25.1 & 62.0 & 42.5 & 48.8  \\
            +\methodname & & \textbf{41.2} & \textbf{73.0} & \textbf{51.5} & \textbf{61.4}   \\
            \midrule
            {Hymba} & \multirow{2}{*}{15} & 21.8 & 60.0 & 41.5 & 48.0\\
            +\methodname & & \textbf{37.1} & \textbf{70.0} & \textbf{48.9} & \textbf{57.7} \\
            \midrule
            {Hymba} & \multirow{2}{*}{23} & 5.9 & 41.0 & 27.8 & 32.3  \\
            +\methodname & & \textbf{29.1} & \textbf{62.0} & \textbf{42.8} & \textbf{50.1}   \\
            \midrule
            {Hymba} & \multirow{2}{*}{29} & 5.8 & 40.0 & 27.6 & 30.3  \\
            +\methodname & & \textbf{31.2} & \textbf{57.0} & \textbf{39.2} & \textbf{44.5}   \\
            \bottomrule
        \end{tabular}
    }
\end{wraptable}
Rows 5 and 6 in \cref{tab:qa} shows these results on a Hymba model, again on the tested QA datasets. The \methodname-augmented variant remains significantly more performant than the other alternatives, demonstrating the benefits that acting directly on the representations can have.

\paragraph{Relationship with Hybrid Methods.}\label{subsec:analysis}
Some previous works have suggested methods that enable linear recurrent models to improve their in-context learning abilities. \citet{mamba-icl} introduce a {MambaFormer} architecture which interleaves self-attention and {Mamba} layers, improving the ability to learn in-context on tasks in which pure {Mamba} models struggle. Similarly, \methodname~can be interpreted as a form of hybrid mixture of attention and recurrence, similar to \citet{hymba}, with the difference lying in the frequency of attention and its sparsity within different layers. To better compare these two methods, we provide an ablation (\cref{tab:qa-multidepth}) where we replace Hymba layers with \methodname. In this setting, 3 consecutive layers are trained in either setup. We observe a meaningful increase in performance on \textsc{CoQA}, indicating that for such types of tasks, the mechanism presented by \methodname~could be more robust and suitable on a number of real world tasks.

\paragraph{Ablating on the position of layers.} Due to its nature, a natural question that emerges relates to the ease and effectiveness of determining the layers at which \methodname~augmentations need to be made. The same results~(\cref{tab:qa-multidepth}) show that the placement of the 3 \methodname~layers do not have a significant impact on the performance improvement relative to the baseline, which can be improved upon in nearly all ways in which the layers are selected. This highlights a level of robustness of the framework and method, hinting towards an ability to be used for a variety of additional tasks.

\paragraph{Efficiency Evaluation.} Given the architectural modifications that following from the \methodname~augmentations, a comparison between the efficiency tradeoffs is also necessary. We provide ablations using \textbf{Mamba}, \textbf{DeltaNet} and \textbf{GLA}, comparing both base models and those augmented with \methodname~to demonstrate these tradeoffs in \cref{tab:prefilling}-\ref{tab:memory}. While the \methodname~augmentations do lead to an increase in each factor, the plug-in remains lightweight and does not add significant overhead in computation, particularly for longer sequences. Furthermore, direct comparison with a Transformer shows this method to be significantly more lightweight while retaining Transformer-like performance on our tasks. 

\begin{table}[h]
    \centering
    \resizebox{\linewidth}{!}{
    \begin{tabular}{c|c|cc|cc|cc}
        \toprule
        \textbf{Pre-filling Length} & \textbf{Transformer} & \textbf{Mamba} & \textbf{Mamba + Resona} & \textbf{DeltaNet} & \textbf{DeltaNet + Resona} & \textbf{GLA} & \textbf{GLA + Resona} \\
        \midrule
        \midrule
        2k  & 29  & 45  & 52  & 53  & 64  & 37  & 46  \\
        4k  & 34  & 77  & 88  & 59  & 78  & 43  & 64  \\
        8k  & 71  & 149 & 170 & 72  & 103 & 62  & 104 \\
        16k & 173 & 294 & 349 & 106 & 181 & 109 & 202 \\
        32k & 503 & 571 & 653 & 208 & 338 & 202 & 386 \\
        64k & 1665 & 1118 & 1285 & 412 & 652 & 407 & 757 \\
        128k & 6094 & 2257 & 2412 & 807 & 1289 & 806 & 1518 \\
        \bottomrule
    \end{tabular}
    }
    \caption{The time (in milliseconds) used to pre-fill a context of a pre-specified length. Numbers are rounded to the nearest millisecond.}
    \label{tab:prefilling}
\end{table}

\begin{table}[h]
    \centering
    \resizebox{\linewidth}{!}{
    \begin{tabular}{c|c|cc|cc|cc}
        \toprule
        \textbf{Pre-filling Length} & \textbf{Transformer} & \textbf{Mamba} & \textbf{Mamba + Resona} & \textbf{DeltaNet} & \textbf{DeltaNet + Resona} & \textbf{GLA} & \textbf{GLA + Resona} \\
        \midrule
        \midrule
        2k  & 2679 & 2972 & 3329 & 2945 & 3686 & 2749 & 3528 \\
        4k  & 2758 & 3044 & 3360 & 3042 & 3772 & 2777 & 3543 \\
        8k  & 2866 & 3155 & 3499 & 3023 & 3789 & 2774 & 3523 \\
        16k & 3389 & 3164 & 3613 & 3057 & 3829 & 2869 & 3689 \\
        32k & 5759 & 3491 & 3910 & 3080 & 4207 & 2912 & 4044 \\
        64k & 11144 & 4119 & 4656 & 3171 & 4404 & 3145 & 4352 \\
        128k & 24050 & 4747 & 5601 & 3611 & 5278 & 3509 & 5076 \\
        \bottomrule
    \end{tabular}
    }
    \caption{The time (in milliseconds) taken to generate 128 tokens following a prespecified pre-filling length. Numbers are rounded to the nearest millisecond.}
    \label{tab:generation}
\end{table}

\begin{table}[h!]
    \centering
    \resizebox{\linewidth}{!}{
    \begin{tabular}{c|c|cc|cc|cc}
        \toprule
        \textbf{Pre-filling Length} & \textbf{Transformer} & \textbf{Mamba} & \textbf{Mamba + Resona} & \textbf{DeltaNet} & \textbf{DeltaNet + Resona} & \textbf{GLA} & \textbf{GLA + Resona} \\
        \midrule
        \midrule
        2k  & 3.1  & 2.8  & 4.0  & 2.9  & 4.8  & 2.9  & 3.6  \\
        4k  & 3.5  & 2.8  & 4.1  & 3.0  & 4.9  & 3.0  & 3.7  \\
        8k  & 4.5  & 3.1  & 4.4  & 3.3  & 5.2  & 3.2  & 3.9  \\
        16k & 6.4  & 3.7  & 5.1  & 3.7  & 5.5  & 3.6  & 4.4  \\
        32k & 10.2 & 4.9  & 6.5  & 4.6  & 6.6  & 4.3  & 5.4  \\
        64k & 17.7 & 7.2  & 9.2  & 6.4  & 8.4  & 5.9  & 7.5  \\
        128k & 32.9 & 11.9 & 14.7 & 10.2 & 12.3 & 9.1  & 11.7 \\
        \bottomrule
    \end{tabular}
    }
    \caption{The total memory consumption (in GB) given a pre-specified pre-filling length.}
    \label{tab:memory}
\end{table}

\section{Conclusion}

In this work, we propose \methodname, a lightweight retrieval-based knowledge integration mechanism that significantly improves the ability of LRMs to use example-specific context. \methodname~utilizes a novel mechanism to use the input-context as a retrieval data-store and integrate such information with the input during training and inference, enabling models to use it more effectively and overcome information bottleneck concerns. Across a number of both synthetic and real-world datasets, LRMs augmented with \methodname~demonstrate significant performance gains compared to their base counterparts, demonstrating its ability to function as a general method applicable to broader scenarios.

\clearpage
\bibliography{refs}
\bibliographystyle{abbrvnat}

\newpage
\appendix
\onecolumn

\section{Additional Results}

\subsection{Detailed Pretraining Results}

\begin{table*}[ht]
    \centering
    \caption{Detailed \texttt{lm-evaluation-harness} evaluation results from pre-training.}
    \resizebox{\linewidth}{!}{
        \begin{tabular}{l|c|c|cc|cc|cc|cc|cc|c|c|c|c}
            \toprule
            & \textbf{Param} & \textbf{Wiki.} & \multicolumn{2}{c|}{\textbf{ARC-C}} & \multicolumn{2}{c|}{\textbf{ARC-E}} & \multicolumn{2}{c|}{\textbf{Hella.}} & \multicolumn{2}{c|}{\textbf{OBQA}} & \multicolumn{2}{c|}{\textbf{PIQA}} & \textbf{PM} & \textbf{RACE} & \textbf{Wino.} & \textbf{AVG} \\
            & & \textsf{ppl} & \textsf{acc} & \textsf{acc\_n} & \textsf{acc} & \textsf{acc\_n} & \textsf{acc} & \textsf{acc\_n} & \textsf{acc} & \textsf{acc\_n} & \textsf{acc} & \textsf{acc\_n} & \textsf{acc} & \textsf{acc} & \textsf{acc} & \textsf{acc} \\
            \midrule
            \midrule
            {GLA} & 131M & 14.265 & 0.1843 & 0.221 & 0.3683 & 0.3502 & 0.2671 & 0.2692 & 0.184 & 0.268 & 0.5392 & 0.5109 & 0.336 & 0.2364 & 0.5028 & 0.3259 \\
            +(SP) & 142M & 14.223 & 0.2073 & 0.2483 & 0.3409 & 0.3418 & 0.2701 & 0.2696 & 0.158 & 0.234 & 0.5321 & 0.5141 & 0.358 & 0.2478 & 0.4949 & 0.3243 \\
            +\methodname & 145M & \textbf{13.892} & 0.192 & 0.2517 & 0.3405 & 0.3253 & 0.2646 & 0.2716 & 0.174 & 0.272 & 0.5457 & 0.5261 & 0.352 & 0.2469 & 0.5249 & \textbf{0.3297} \\
            \midrule
            {DeltaNet} & 131M & 13.044 & 0.2201 & 0.2628 & 0.2849 & 0.2912 & 0.2613 & 0.2607 & 0.146 & 0.262 & 0.5234 & 0.5038 & 0.35 & 0.2545 & 0.5059 & 0.3174 \\
            +(SP) & 142M & 12.946 & 0.2159 & 0.256 & 0.3001 & 0.2845 & 0.2592 & 0.2655 & 0.178 & 0.278 & 0.5305 & 0.5125 & 0.34 & 0.266 & 0.4728 & 0.3199 \\
            +\methodname & 145M & \textbf{12.541} & 0.2073 & 0.2449 & 0.2963 & 0.2963 & 0.2686 & 0.2757 & 0.156 & 0.272 & 0.5419 & 0.5283 & 0.334 & 0.2517 & 0.5012 & \textbf{0.3210} \\
         
            \bottomrule
        \end{tabular}
    }
    \label{tab:pretraining-detail}
\end{table*}

\subsection{Detailed Supervised Fine-Tuning Results}
\begin{table*}[h!]
    \centering
    \caption{Results on Needle-in-a-Haystack (\textsc{NIAH}) using a haystack of varying sizes. Models are scored on performance on a continuous scale from 0 (worst) to 5 (best). In all settings, there is a single needle placed arbitrarily within the haystack. Different variants mean that the format of the needle or haystack changes, such as being a number, keyword or UUID sequence. Here $\alpha\times \mathsf{lr}$ denotes \methodname~is trained with a learning rate multiplied by $\alpha$.}
    \resizebox{\linewidth}{!}{
        \begin{tabular}{lc|cccc|cccc|cccc}
            \toprule
            \multirow{2}{*}{\textbf{Model}} & \multirow{2}{*}{\textbf{Setting}} & \multicolumn{4}{c|}{{4$\mathsf{K}$}} & \multicolumn{4}{c|}{{8$\mathsf{K}$}} & \multicolumn{4}{c}{{16$\mathsf{K}$}} \\
            & & V1  & V2 & V3 & MV & V1 & V2 & V3 & MV & V1  & V2 & V3 & MV \\
            \midrule
            \midrule
            \multirow{3}{*}{Mamba} & \texttt{Baseline} & 5.00 & 1.80 & 0.75 & 0.898 & 0.65 & 0.45 & 0.15 & 0.458 & 0.00 & \textbf{0.35} & \textbf{0.20} & 0.494 \\
            & 20$\times \mathsf{lr}$ & 5.00 & \textbf{3.65} & \textbf{1.80} & \textbf{1.358} & \textbf{5.00} & \textbf{0.85} & \textbf{0.45} & 0.528 & \textbf{4.90} & 0.05 & 0.00 & \textbf{0.669} \\
            & 50$\times \mathsf{lr}$ & 5.00 & 3.00 & 1.10 & 1.267 & 4.95 & 0.60 & 0.20 & \textbf{0.533} & 4.00 & 0.05 & 0.00 & 0.550  \\
            \midrule
            \multirow{3}{*}{DeltaNet} & \texttt{Baseline} & 2.15 & 2.95 & 1.45 & 1.301 & 1.90 & 1.45 & 0.70 & 0.321 & 1.00 & 0.25 & 0.00 & 0.883 \\
            & 20$\times \mathsf{lr}$ & 5.00 & 4.10 & 0.60 & 0.876 & 5.00 & \textbf{2.60} & \textbf{0.80} & \textbf{1.453} & 5.00 & \textbf{0.60} & \textbf{0.10} & \textbf{0.973} \\
            & 50$\times \mathsf{lr}$ & \textbf{5.00} & \textbf{4.40} & \textbf{1.45} & \textbf{1.676 }& \textbf{5.00} & 1.25 & 0.35 & 1.312 & \textbf{5.00} & 0.45 & 0.05 & 0.885 \\
            \midrule
            \multirow{3}{*}{GLA} & \texttt{Baseline} & 2.90 & 3.65 & 0.05 & \textbf{1.389} & 0.25 & 0.55 & 0.00 & 0.846 & 0.00 & 0.05 & 0.00 & 0.291 \\
            & 20$\times \mathsf{lr}$ & 2.55 & \textbf{3.90} & 0.10 & 1.310 & \textbf{0.30} & 0.50 & 0.00 & 0.930 & 0.00 & \textbf{0.10} & 0.00 & 0.578 \\
            & 50$\times \mathsf{lr}$ & \textbf{3.00} & 3.85 & \textbf{0.15} & 1.353 & 0.20 &\textbf{ 0.60} & 0.00 & \textbf{0.900} & 0.00 & 0.05 & 0.00 & \textbf{0.657} \\
            \bottomrule
        \end{tabular}
    }

    \label{tab:niah}
\end{table*}

\begin{table*}[h!]
    \centering
    
    \caption{Detailed \texttt{lm-evaluation-harness} evaluation results from supervised fine-tuning of different pre-trained LRMs. After undergoing the same SFT as the backbone models, \methodname-enhanced models achieve comparable or superior zero-shot lm-harness evaluation scores to baselines. Combined with \cref{tab:niah}, these results demonstrate that the \methodname module enhances the backbone's in-context learning (ICL) capability while maintaining its foundational language modeling performance.}
    \resizebox{\linewidth}{!}{
        \begin{tabular}{l|cc|cc|cc|cc|cc|c|c|c|c}
            \toprule
            & \multicolumn{2}{c|}{\textbf{ARC-C}} & \multicolumn{2}{c|}{\textbf{ARC-E}} & \multicolumn{2}{c|}{\textbf{Hella.}} & \multicolumn{2}{c|}{\textbf{OBQA}} & \multicolumn{2}{c|}{\textbf{PIQA}} & \textbf{PM} & \textbf{RACE} & \textbf{Wino.} & \textbf{AVG} \\
            & \textsf{acc} & \textsf{acc\_n} & \textsf{acc} & \textsf{acc\_n} & \textsf{acc} & \textsf{acc\_n} & \textsf{acc} & \textsf{acc\_n} & \textsf{acc} & \textsf{acc\_n} & \textsf{acc} & \textsf{acc} & \textsf{acc} & \textsf{acc} \\
            \midrule
            \midrule
            {GLA}  & 0.2381 & 0.2747 & 0.5442 & 0.5046 & 0.3852 & 0.4903 & 0.198 & 0.314 & 0.7008 & 0.7008 & 0.552 & 0.3110 & 0.5288 & 0.4417 \\
            +\methodname & 0.2466 & 0.2918 & 0.5497 & 0.5227 & 0.3738 & 0.4722 & 0.188 & 0.308 & 0.6944 & 0.7010 & 0.550 & 0.3139 & 0.5399 & \textbf{0.4424} \\
            \midrule
            {DeltaNet} & 0.2363 & 0.2637 & 0.5636 & 0.5341 & 0.3852 & 0.4893 & 0.198 & 0.316 & 0.7035 & 0.7002 & 0.552 & 0.3388 & 0.5375 & 0.4475 \\
            +\methodname & 0.2440 & 0.2722 & 0.5812 & 0.5455 & 0.3909 & 0.4982 & 0.198 & 0.314 & 0.7024 & 0.6997 & 0.556 & 0.3292 & 0.5375 & \textbf{0.4514} \\
            \midrule
            {Mamba} & 0.3558 & 0.3805 & 0.6953 & 0.6427 & 0.4958 & 0.6490 & 0.270 & 0.382 & 0.7535 & 0.7535 & 0.684 & 0.3598 & 0.6440 & 0.5435 \\
            +\methodname & 0.3823 & 0.3951 & 0.6907 & 0.6904 & 0.4804 & 0.6292 & 0.300 & 0.406 & 0.7372 & 0.7383 & 0.690 & 0.3445 & 0.6338 & \textbf{0.5475} \\

            \bottomrule
        \end{tabular}
    }
    \label{tab:finetuning-detail}
\end{table*}

\subsection{Training Details for Synthetic Benchmarks}

\paragraph{Multi-Query Associative Recall (\textsc{MQAR}).}
We evaluate \methodname~on the \textsc{MQAR} task by training six base architectures: \textbf{BaseConv}, \textbf{H3}, \textbf{Hyena}, \textbf{Mamba}, \textbf{Mamba2}, and \textbf{RWKV}. All models are trained with a 4-layer configuration and a hidden dimension (\textit{d\_model}) ranging from 32 to 256. The sequence lengths vary from 64 to 512, with the number of key-value (KV) pairs corresponding to 4--32, respectively. To integrate \methodname, we insert the \textbf{Resona Layer} into either the first or third layer of each model, using the same \textit{d\_model} settings. The learning rate is swept using the default settings from \citet{zoology_mqar}. For the \textbf{Resona Layer}, we use a chunk size of 2 and a top-$k$ value of 1. The exact results for each model, sequence length, and hidden dimension can be found in \cref{tab:accuracy}.
\begin{table*}[ht]
    \centering
    \caption{Best Accuracy for Different Models with Varying Sequence Length and Model Dimensions}
    \setlength{\tabcolsep}{4pt}
    \resizebox{\linewidth}{!}{
    \begin{tabular}{l|cccc|cccc|cccc|cccc}
        \toprule
        \multirow{3}{*}{\textbf{Model}} & \multicolumn{4}{c|}{$L=64$} & \multicolumn{4}{c|}{$L=128$} & \multicolumn{4}{c|}{$L=256$} & \multicolumn{4}{c}{$L=512$} \\
        \cmidrule(lr){2-5} \cmidrule(lr){6-9} \cmidrule(lr){10-13} \cmidrule(lr){14-17}
        & \multicolumn{4}{c|}{KV Pairs} & \multicolumn{4}{c|}{KV Pairs} & \multicolumn{4}{c|}{KV Pairs} & \multicolumn{4}{c}{KV Pairs} \\
        & 32 & 64 & 128 & 256 & 32 & 64 & 128 & 256 & 32 & 64 & 128 & 256 & 32 & 64 & 128 & 256 \\
        \midrule
            \midrule
        {BaseConv} & 0.401 & 0.439 & 0.805 & 0.960 & 0.051 & 0.115 & 0.469 & 0.948 & 0.017 & 0.046 & 0.076 & 0.549 & 0.003 & 0.007 & 0.015 & 0.023 \\
        +\methodname 0 & 1.000 & 1.000 & 1.000 & 1.000 & 1.000 & 1.000 & 1.000 & 1.000 & 1.000 & 1.000 & 1.000 & 1.000 & 1.000 & 1.000 & 1.000 & 1.000 \\
        +\methodname 2 & 1.000 & 1.000 & 1.000 & 1.000 & 1.000 & 1.000 & 1.000 & 1.000 & 1.000 & 1.000 & 1.000 & 1.000 & 1.000 & 1.000 & 1.000 & 1.000 \\
        \midrule
        {H3} & 0.861 & 0.892 & 0.991 & 0.991 & 0.225 & 0.477 & 0.974 & 0.965 & 0.094 & 0.409 & 0.303 & 0.896 & 0.003 & 0.007 & 0.038 & 0.078 \\
        +\methodname 0 & 1.000 & 1.000 & 1.000 & 1.000 & 1.000 & 1.000 & 1.000 & 1.000 & 1.000 & 1.000 & 1.000 & 1.000 & 1.000 & 1.000 & 1.000 & 1.000 \\
        +\methodname 2 & 1.000 & 1.000 & 1.000 & 1.000 & 1.000 & 1.000 & 1.000 & 1.000 & 1.000 & 1.000 & 1.000 & 1.000 & 1.000 & 1.000 & 1.000 & 1.000 \\
        \midrule
        {Hyena} & 0.430 & 0.874 & 0.922 & 0.972 & 0.051 & 0.501 & 0.646 & 0.840 & 0.018 & 0.089 & 0.313 & 0.928 & 0.002 & 0.007 & 0.053 & 0.173 \\
        +\methodname 0 & 1.000 & 1.000 & 1.000 & 1.000 & 1.000 & 1.000 & 1.000 & 1.000 & 1.000 & 1.000 & 1.000 & 1.000 & 1.000 & 1.000 & 1.000 & 1.000 \\
        +\methodname 2 & 1.000 & 1.000 & 1.000 & 1.000 & 1.000 & 1.000 & 1.000 & 1.000 & 1.000 & 1.000 & 1.000 & 1.000 & 1.000 & 1.000 & 1.000 & 1.000 \\
        \midrule
        {RWKV} & 0.727 & 0.955 & 0.992 & 0.953 & 0.216 & 0.451 & 0.807 & 0.986 & 0.066 & 0.093 & 0.435 & 0.820 & 0.001 & 0.002 & 0.005 & 0.010 \\
        +\methodname 0 & 1.000 & 1.000 & 1.000 & 1.000 & 1.000 & 1.000 & 1.000 & 1.000 & 1.000 & 1.000 & 1.000 & 1.000 & 1.000 & 1.000 & 1.000 & 1.000 \\
        +\methodname 2 & 1.000 & 1.000 & 1.000 & 1.000 & 1.000 & 1.000 & 1.000 & 1.000 & 1.000 & 1.000 & 1.000 & 1.000 & 1.000 & 1.000 & 1.000 & 1.000 \\
        \midrule
        {Mamba} & 0.987 & 0.992 & 0.975 & 0.990 & 0.906 & 0.992 & 0.991 & 0.920 & 0.864 & 0.990 & 0.990 & 0.991 & 0.000 & 0.419 & 0.968 & 0.000 \\
        +\methodname 0 & 1.000 & 1.000 & 1.000 & 1.000 & 1.000 & 1.000 & 1.000 & 1.000 & 1.000 & 1.000 & 1.000 & 1.000 & 1.000 & 1.000 & 1.000 & 1.000 \\
        +\methodname 2 & 1.000 & 1.000 & 1.000 & 1.000 & 1.000 & 1.000 & 1.000 & 1.000 & 1.000 & 1.000 & 1.000 & 1.000 & 1.000 & 1.000 & 1.000 & 1.000 \\
        \midrule
        {Mamba2} & 0.990 & 0.992 & 0.991 & 0.993 & 0.974 & 0.969 & 0.992 & 0.991 & 0.755 & 0.000 & 0.991 & 0.592 & 0.000 & 0.000 & 0.001 & 0.000 \\
        +\methodname 0 & 1.000 & 1.000 & 1.000 & 1.000 & 1.000 & 1.000 & 1.000 & 1.000 & 1.000 & 1.000 & 1.000 & 1.000 & 1.000 & 1.000 & 1.000 & 1.000 \\
        +\methodname 2 & 1.000 & 1.000 & 1.000 & 1.000 & 1.000 & 1.000 & 1.000 & 1.000 & 1.000 & 1.000 & 1.000 & 1.000 & 1.000 & 1.000 & 1.000 & 1.000 \\
        \bottomrule
        \end{tabular}
    }
    \label{tab:accuracy}
\end{table*}

\paragraph{Mechanistic Architecture Design (\textsc{MAD}) Suite.}

For the \textsc{MAD} tasks, we adopt a 4-hybrid block configuration, where each block consists of a linear recurrent layer followed by a SwiGLU layer. We follow the benchmark settings from \citet{madlab}, using a batch size of 128 and a learning rate range of $1\text{e-}3$ to $1\text{e-}4$. For the \methodname~layers, we set the chunk size to 6 and the top-$k$ value to 1.

\subsection{Training Details for Real-World Tasks}

\paragraph{Pretraining.} During pre-training, we prioritized maintaining consistent model depth across architectures. For \texttt{GLA}, \texttt{DeltaNet}, and \texttt{RetNet}, we adopted the architecture implementations from \citet{fla}, configuring them with 24 layers and a hidden size of 600. For \texttt{Hymba}, we use NVIDIA's official 150M parameter implementation (24 layers with a hidden size of 512). For Mamba, we employed FLA-Hub's implementation with 48 layers and a hidden size of 600. For comparative models with equivalent parameter counts, we adjusted the hidden size from 600 to 640. Following the methodology outlined in the \texttt{Hymba} paper, we integrated \methodname modules at the shallowest, middle, and deepest layers to reinforce critical information flow. Each model was trained for 8,000 steps, with model selection performed using a dedicated validation set. The training configuration employed a cosine annealing scheduler with warmup over 5\% of the training steps, the AdamW optimizer (learning rate of $\mathsf{1e\text{-}3}$ and weight decay = $\mathsf{0.01}$), and gradient clipping of $\mathsf{1.0}$.

\paragraph{Finetuning.} We employed a base learning rate of $\mathsf{1e\text{-}5}$ for fine-tuning on the three individual datasets, with other training configurations remaining similar to those used in pre-training, except that \methodname modules benefited from a higher learning rate. For CoQA, NarrativeQA, and TriviaQA, we trained for 2\textsf{K}, 8\textsf{K}, and 10\textsf{K} steps, respectively. During general supervised fine-tuning, we created a unified training set by shuffling 10\textsf{K} samples from each of the three QA datasets, ensuring significant diversity in sequence length and content. Additionally, we constructed a general test set by selecting 500 validation samples from each dataset. We trained on this combined dataset for 4\textsf{K} steps.

\subsection{NIAH Scoring Details}

For the NIAH (Needle In A Haystack) task, we employ an automated scoring protocol based on prefix matching. The scoring methodology operates as follows: A full score of 5 points is awarded if the model's response contains the complete and accurate needle. If not, we iteratively truncate the needle from the end (removing the last few characters incrementally) and perform prefix matching. Partial credit (a proportional fraction of the 5-point maximum) is granted when a truncated prefix matches exactly, with the remaining character percentage determining the awarded score. Responses containing no matching prefix of the needle receive 0 points. The final task score is obtained by averaging the scores across 100 test samples.

\end{document}

%% file: math_commands.tex
\usepackage{amsmath,amsfonts,bm}

\def\eqref#1{equation~\ref{#1}}

\def\1{\bm{1}}

\def\vh{{\bm{h}}}

\def\vs{{\bm{s}}}

\def\vx{{\bm{x}}}
\def\vy{{\bm{y}}}

\def\mC{{\bm{C}}}

\def\mE{{\bm{E}}}

\def\mH{{\bm{H}}}

\def\mK{{\bm{K}}}

\def\mM{{\bm{M}}}

\def\mO{{\bm{O}}}

\def\mQ{{\bm{Q}}}

\def\mV{{\bm{V}}}

\def\mX{{\bm{X}}}
\def\mY{{\bm{Y}}}

\DeclareMathAlphabet{\mathsfit}{\encodingdefault}{\sfdefault}{m}{sl}
\SetMathAlphabet{\mathsfit}{bold}{\encodingdefault}{\sfdefault}{bx}{n}
\newcommand{\tens}[1]{\bm{\mathsfit{#1}}}

\def\tH{{\tens{H}}}

\def\tX{{\tens{X}}}

